# Computing Probability Intervals Under Independency Constraints

Linda van der Gaag
Utrecht University, Department of Computer Science
P.O. Box 80.089, 3508 TB Utrecht, The Netherlands

## Abstract

Many AI researchers argue that probability theory is only capable of dealing with uncertainty in situations where a fully specified joint probability distribution is available, and conclude that it is not suitable for application in AI systems. Probability intervals, however, constitute a means for expressing incompleteness of information. We present a method for computing probability intervals for probabilities of interest from a partial specification of a joint probability distribution. Our method improves on earlier approaches by allowing for independency relationships between statistical variables to be exploited.

## 1 Introduction

The adversaries of probability theory for dealing with uncertainty in AI systems often argue that it is not expressive enough to cope with the different kinds of uncertainty that are encountered in real-life situations. More in specific, it has been argued that probability theory is not able to distinguish between uncertainty and ignorance due to incompleteness of information. The suitability of probability intervals for expressing incompleteness has been pointed out decisively by J. Pearl in [Pearl, 1988a]. In this paper, we present a framework for computing probability intervals from an incomplete set of probabilities. The general idea of our approach is to take the initially given probabilities as defining constraints on a yet unknown joint probability distribution. Several authors have already addressed the problem of computing probability intervals, see for example [Cooper, 1984] and [Nilsson, 1986]. Our approach differs from the mentioned ones by taking independency relationships between the statistical variables discerned into consideration. In order to do so, we assume that the independencies in the unknown distribution are specified in a special type of graph. The topological properties of this graph are then exploited for computing precise intervals.

In Section 2 we introduce the notion of a partial specification of a joint probability distribution. Furthermore, the foundation for our method for computing probability intervals from such a partial specification is layed. In Section 3 we discuss how independency constraints can be taken into consideration. In Section 4 we briefly point out that our method can be of real help in the process of knowledge acquisition for so-called belief networks.

## 2 Computing Probability Intervals

In this section, we concentrate ourselves on the notion of a partial specification of a joint probability distribution and develop a method for computing probability intervals for probabilities of interest from such a partial specification. For the moment, we assume that no independencies between the statistical variables discerned exist. We begin by introducing some terminology.

Let $\mathcal{B}(a_1, \ldots, a_n)$ be a free Boolean algebra generated by a set of atomic propositions $\mathcal{A} = \{a_1, \ldots, a_n\}$, $n \geq 1$; alternatively, the Boolean algebra $\mathcal{B}(a_1, \ldots, a_n)$ may be viewed as a sample space being 'spanned' by a set of statistical variables $A_i$, $i = 1, \ldots, n$, each taking values from $\{a_i, \neg a_i\}$. A *partial specification of a joint probability distribution* on $\mathcal{B}(a_1, \ldots, a_n)$ is a total function $P : \mathcal{C} \to [0, 1]$ where $\mathcal{C} \subseteq \mathcal{B}(a_1, \ldots, a_n)$. We call such a partial specification *consistent* if there exists at least one joint probability distribution $Pr$ on $\mathcal{B}$ such that $Pr$ is an extension of $P$ onto $\mathcal{B}(a_1, \ldots, a_n)$ (notation: $Pr|_\mathcal{C} = P$); otherwise, $P$ is said to be *inconsistent*. Furthermore, we say that $P$ (uniquely) *defines* $Pr$ if $Pr$ is the only joint probability distribution on $\mathcal{B}(a_1, \ldots, a_n)$ such that $Pr|_\mathcal{C} = P$. Now, let $\mathcal{B}_0$ be the subset of $\mathcal{B}(a_1,\ldots,a_n)$ consisting of its 'smallest elements', that is, let

$$\mathcal{B}_0 = \{\bigwedge_{i=1}^n L_i \mid L_i = a_i \text{ or } L_i = \neg a_i, a_i \in \mathcal{A}\}$$

We state some convenient properties of this set $\mathcal{B}_0$. Let the elements of $\mathcal{B}_0$ be enumerated as $b_i$, $i = 1,\ldots,2^n$. Then, for any joint probability distribution $Pr$ on $\mathcal{B}(a_1,\ldots,a_n)$ we have that

$$\sum_{i=1}^{2^n} Pr(b_i) = 1 \qquad (1)$$

The probabilities $Pr(b_i)$ for the elements $b_i \in \mathcal{B}_0$ will be called the *constituent probabilities* of $Pr$. Furthermore, for each element $b \in \mathcal{B}(a_1,\ldots,a_n)$ there exists a unique set of indices $\mathcal{I}_b \in \{1,\ldots,2^n\}$ such that $b = \bigvee_{i \in \mathcal{I}_b} b_i$; this set $\mathcal{I}_b$ will be called the *index set* for $b$. For each joint probability distribution $Pr$ on $\mathcal{B}(a_1,\ldots,a_n)$ we then have that

$$Pr(b) = \sum_{i \in \mathcal{I}_b} Pr(b_i) \qquad (2)$$

In addition, it can easily be shown that any consistent partial specification $P : \mathcal{B}_0 \to [0,1]$ defined on $\mathcal{B}_0$ uniquely defines a joint probability distribution $Pr$ on the entire algebra $\mathcal{B}(a_1,\ldots,a_n)$. In the sequel, we will write $\mathcal{B}$ instead of $\mathcal{B}(a_1,\ldots,a_n)$ as long as ambiguity cannot occur.

We exploit the set $\mathcal{B}_0$ and its properties for computing probability intervals from an arbitrary partial specification. Suppose that we are given probabilities for a number of arbitrary elements of the Boolean algebra $\mathcal{B}$, that is, we consider the case in which we are given a consistent partial specification $P$ of a joint probability distribution on $\mathcal{B}$ that is defined on an arbitrary subset $C \subseteq \mathcal{B}$. The problem of computing probability intervals from $P$ will now be transformed into an equivalent problem in linear algebra. The general idea is to take the initially given probabilities as defining constraints on a yet unknown joint probability distribution.

Let $C = \{c_1,\ldots,c_m\}$, $m \geq 1$, be a subset of $\mathcal{B}$, and let $P : C \to [0,1]$ be a consistent partial specification of a joint probability distribution on $\mathcal{B}$. We now consider an arbitrary (yet unknown) joint probability distribution $Pr$ on $\mathcal{B}$ with $Pr|_C = P$. Let the constituent probabilities $Pr(b_i)$, $b_i \in \mathcal{B}_0$, of $Pr$ be denoted by $x_i$, $i = 1,\ldots,2^n$, and let the initially specified probabilities $P(c_i) = Pr(c_i), c_i \in C, i = 1,\ldots,m$, be denoted by $p_i$. Using (1) and (2), we obtain the following inhomogeneous system of linear equations:

$$\begin{array}{ccccccc}
d_{1,1}x_1 & + & \ldots & + & d_{1,2^n}x_{2^n} & = & p_1 \\
\vdots & & & & \vdots & & \vdots \\
d_{m,1}x_1 & + & \ldots & + & d_{m,2^n}x_{2^n} & = & p_m \\
x_1 & + & \ldots & + & x_{2^n} & = & 1
\end{array}$$

where $d_{i,j} = \begin{cases} 0 & \text{if } j \notin \mathcal{I}_{c_i} \\ 1 & \text{if } j \in \mathcal{I}_{c_i} \end{cases}$, $i = 1,\ldots,m$, $j = 1,\ldots,2^n$, in which $\mathcal{I}_{c_i}$ is the index set for $c_i \in C$. This system of linear equations has the $2^n$ unknowns $x_1,\ldots,x_{2^n}$. Now, let $p$ denote the column vector of right-hand sides and $x$ the vector of unknowns. Furthermore, let $D$ denote the coefficient matrix of the system. We will use the matrix equation $Dx = p$ to denote the system of linear equations obtained from a partial specification $P$ as described above.

We have the following relation between extensions of a consistent partial specification of a joint probability distribution and solutions to the matrix equation obtained from it:

- For any joint probability distribution $Pr$ on $\mathcal{B}$ such that $Pr|_C = P$, we have that the vector $x$ of constituent probabilities $x_i = Pr(b_i)$, $b_i \in \mathcal{B}_0$, $i = 1,\ldots,2^n$, is a solution to $Dx = p$.

- For any nonnegative solution vector $x$ with components $x_i$, $i = 1,\ldots,2^n$, to $Dx = p$, we have that $Pr(b_i) = x_i$, $b_i \in \mathcal{B}_0$, defines a joint probability distribution $Pr$ on $\mathcal{B}$ such that $Pr|_C = P$.

Note that although every joint probability distribution $Pr$ which is an extension of $P$ corresponds uniquely with a solution to the matrix equation $Dx = p$ obtained from $P$, not every solution to $Dx = p$ corresponds with a 'probabilistic' extension of $P$: $Dx = p$ may have solutions in which at least one of the $x_i$'s is less than zero.

It can easily be shown that the problem of finding for a given $b \in \mathcal{B}$ the least upper bound to the probability of $b$ relative to a partial specification $P$ is equivalent to the following linear programming problem:

$$maximize$$
$$\sum_{j=1}^{2^n} c_j x_j (= Pr(b))$$

$$subject\ to$$

(i) $\sum_{j=1}^{2^n} d_{i,j} x_j = p_i$, for $i = 1,\ldots,|C|+1$,

(ii) $x_j \geq 0$, for $j = 1,\ldots,2^n$

where $c_j = \begin{cases} 0 & \text{if } j \notin \mathcal{I}_b \\ 1 & \text{if } j \in \mathcal{I}_b \end{cases}$ and $d_{i,j}$ constitute the matrix $D$. A similar statement can be made for the greatest lower bound to the probability of $b$ relative to $P$. Note that this linear programming approach can deal with conditional probabilities in the same way in which it handles prior ones; furthermore, the approach allows for initial specifications of bounds to probabilities instead of point estimates.

It is well-known that an LP-problem can be solved in polynomial time, that is, polynomial in the size of the problem, [Papadimitriou, 1982]. The size of an LP-problem is dependent, among other factors, upon the number of variables it comprises. Now note that the specific type of problem discussed in the foregoing has exponentially many variables, that is, exponential in the number of statistical variables discerned in the problem domain. Therefore, computing probability intervals requires an exponential number of steps.

## 3 Exploiting Independency Relationships

In the preceding section we have presented a linear programming method for computing probability intervals from a consistent partial specification of a joint probability distribution. The initially assessed probabilities were viewed as defining constraints on an unknown probability distribution. We assumed that no independency relationships existed between the statistical variables discerned. In this section, the linear programming approach is extended with an additional method for representing and exploiting independency relationships. Note that representing independency relationships in a straightforward manner yields nonlinear constraints and therefore is not suitable for our purposes.

We assume that the independency relationships between the statistical variables have been specified as an *I-map* of the unknown joint probability distribution $Pr$. Informally speaking, an I-map of $Pr$ is an undirected graph in which the vertices represent the statistical variables discerned and the missing edges indicate the independencies holding between the variables. Furthermore, we assume that the fill-in algorithm by R.E. Tarjan and M. Yannakakis has been applied to yield a *decomposable* I-map $G$ of $Pr$. An I-map is decomposable if it does not contain any elementary cycles of length four or more without a shortcut. For further information, the reader is referred to [Pearl, 1988b]. We will show that we can take advantage of the topology of $G$ by observing that between the variables in a clique of $G$ no independency rela-tionships exist and that the cliques are interrelated only through their intersections. In order to be able to exploit these properties, we further assume that all initially given probabilities are local to the cliques of $G$. Once more we introduce some new terminology.

Let $G = (V(G), E(G))$ be a decomposable graph with the vertex set $V(G) = \{V_1, \ldots, V_n\}$, $n \geq 1$, and the clique set $Cl(G) = \{Cl_1, \ldots, Cl_m\}$, $m \geq 1$, to be taken as a decomposable I-map of an unknown joint probability distribution $Pr$. We take the graph from Figure 1(a) as our running example. Let $\mathcal{B}$ be the free Boolean algebra generated by $\{V_i \mid V_i \in V(G)\}$; furthermore, for each clique $Cl_i$, let $\mathcal{B}(Cl_i) \subseteq \mathcal{B}$ be the free Boolean algebra generated by $\{V_j \mid V_j \in V(Cl_i)\}$. Now, let $P$ be a partial specification of a joint probability distribution on $\mathcal{B}$ (recall that all initially given probabilities are local to the cliques of $G$). We say that $P$ is *consistent with respect to $G$* if $P$ can be extended in at least one way to a joint probability distribution $Pr$ on $\mathcal{B}$ such that $Pr$ is decomposable relative to $G$, that is, such that $Pr$ can be expresssed in terms of marginal distributions on the cliques of $G$. The initially given probabilities being local now allows us to apply the notions introduced in the preceding section separately to marginal distributions on the cliques of $G$. We begin by taking the definition of a partial specification of a joint probability distribution to apply to marginal distributions: a *partial specification of a marginal distribution* on $\mathcal{B}(Cl_i)$ is a total function $m_{Cl_i}: C_i \to [0, 1]$ where $C_i \subseteq \mathcal{B}(Cl_i)$. Note that we may now view $P$ as been defined by a set of partial specifications of marginal distributions $M = \{m_{Cl_i} \mid Cl_i \in Cl(G)\}$. Furthermore, we take the notion of consistency to apply to partial specifications of marginal distributions: we call such a partial specification *consistent* if it can be extended in at least one way to an actual marginal distribution.

The analogy between the notions of a consistent partial specification of a joint probability distribution and a consistent partial specification of a marginal distribution suggest that we may apply the linear programming method presented in the preceding section separately to each of the partial specifications of marginal distributions associated with the cliques of $G$. However, even if all partial specifications of marginal distributions have been specified consistently, they might still not give rise to a joint probability distribution that respects the independency relationships shown in $G$. We therefore define some additional notions of consistency:

- The set $M$ of partial specifications of marginal distributions is called *locally consistent* if each $m_{Cl_i} \in M$, $i = 1, \ldots, m$, is consistent.

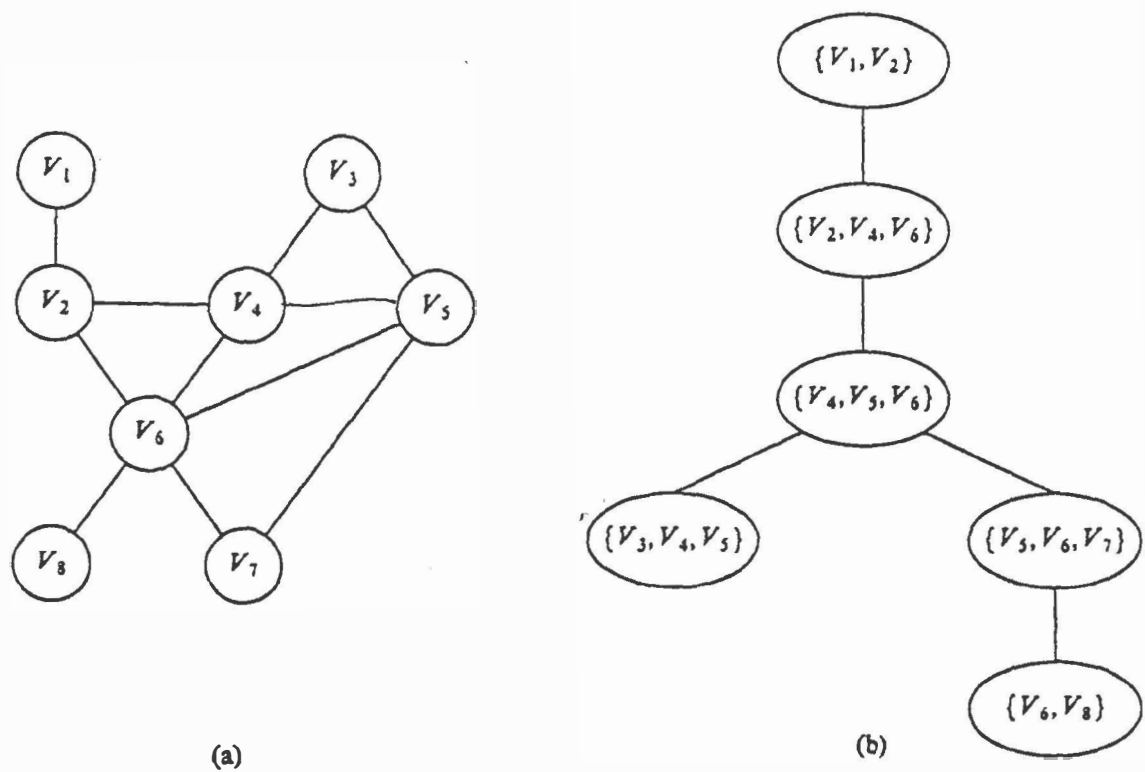

(a)　　　　　　　　　　　　(b)

Figure 1

- $M$ is called *globally consistent* if there exists a set $M = \{\mu_{Cl_i} | \mu_{Cl_i} : B(Cl_i) \to [0,1]\}$ of marginal distributions $\mu_{Cl_i}$ on $B(Cl_i)$ such that for each clique $Cl_i \in Cl(G)$, $\mu_{Cl_i}$ is an extension of $m_{Cl_i} \in M$, and furthermore that for each pair of cliques $Cl_i, Cl_j \in Cl(G)$ with $V(Cl_i) \cap V(Cl_j) \neq \emptyset$ we have that $\mu_{Cl_i}(V(Cl_i) \cap V(Cl_j)) = \mu_{Cl_j}(V(Cl_i) \cap V(Cl_j))$; such a set M is called a *global extension* of $M$.

It can be shown that global consistency of $M$ is a necessary and sufficient condition for $P$ being consistent with respect to $G$; further details are provided in [Gaag, 1990].

We now apply the linear programming method from the preceding section separately to each of the partial specifications of marginal distributions associated with the cliques of $G$. For each clique $Cl_i \in Cl(G)$ we now define a vector $x_i$ of constituent probabilities of a yet unknown marginal distribution $\mu_{Cl_i}$ in the manner described in Section 2. From the partial specification $m_{Cl_i}$ associated with clique $Cl_i$ we then obtain an appropriate system of linear constraints with the constituent probabilities as unknowns. This system will be denoted by $D_i x_i = m_i$, $x_i \geq 0$. The separate systems of constraints are subsequently combined into one large system of linear constraints; this system will be denoted by $Dx = m$, $x \geq 0$. To guarantee that every nonnegative solution to the thus obtained system of constraints defines an extension of the initially given probabilities to a joint probability distribution that is decomposable relative to $G$, we have to augment the system with some additional constraints, called *independency constraints*, expressing that the set $M$ of partial specifications of marginal distributions has to be globally consistent. In theory we now have to obtain for each pair of cliques $Cl_i, Cl_j \in Cl(G)$ with $V(Cl_i) \cap V(Cl_j) \neq \emptyset$, a number of constraints specifying that $\mu_{Cl_i}(V(Cl_i) \cap V(Cl_j)) = \mu_{Cl_j}(V(Cl_i) \cap V(Cl_j))$. However, if we do so, we get many redundant constraints; in fact, the reader may verify that it suffices to obtain independency constraints from the clique intersections represented in a join tree of $G$ only. Figure 1(b) shows a join tree $T_G$ of our example graph. Note that the resulting independency constraints each involve variables from two cliques only. In the sequel, the system of independency constraints for the intersection of two cliques $Cl_i$ and $Cl_j$ will be denoted by $T_{i,j}x_i - T_{j,i}x_j = 0$; the system of independency constraints obtained from an entire clique tree of $G$ will be denoted by $Tx = 0$. From now on we will call $Dx = m, Tx = 0, x \geq 0$, the *joint* system of constraints. Analogous to our observations in Section 2, we have that the problem of finding for a given $b \in \mathcal{B}$ (which is local to a clique of $G$), the least upper bound to the probability of $b$ is equivalent to maximizing the probability of $b$ subject to this joint system of constraints. Again, a similar statement can be made concerning the greatest lower bound to the probability of $b$. It should be evident that in the resulting probability interval the independency relationships shown in $G$ have been taken into account properly.

We can solve the linear programming problem discussed above using a traditional LP-program or a decomposition method like Dantzig-Wolfe decomposition, [Papadimitriou, 1982]. In such a straightforward approach, however, the modular structure of the problem at hand is not fully exploited. We will present an algorithm for solving the problem in which the computations are restricted to local computations per clique only. First, we describe its basic idea informally for our running example.

Consider Figure 2 in which the join tree $T_G$ of $G$ has been depicted once more, this time explicitly showing the clique intersections. We view $T_G$ as a computational architecture in which the vertices are *autonomous objects* holding the local systems of constraints as private data. These objects are only able to communicate with their direct neighbours and only 'through' the independency constraints: the edges are viewed as *communication channels*. The independency constraints are used for relating variables from one clique to variables from another one. Now, suppose that we are interested in the least upper bound to a probability of interest which is local to a specific clique, like the one shown in the figure. The object corresponding with the clique now sends a request for information about further constraints, if any, to its neighbours and then waits until it has received the requested information from all of them. For the moment, each 'interior' object in the join tree just passes the request on to its other neighbours and awaits the requested information. As soon as a leaf (or the root) of the tree receives such a request for information, a second pass through the tree is started. The leaf computes the feasible set of its local system of constraints and derives from it (by means of projection) the set of feasible values for the probabilities which are the constituent probabilities for the intersection(s) with its neighbour(s). This information then is passed on to these neighbours via the appropriate communication channels using the independency constraints for 'translation' of the variables. This results in the addition of extra constraints to the local system of constraints of these neighbours. These computations are performed by the interior vertices as well until the object that started the computation has been reached again. The arcs in Figure 2 represent the flow of computation from this second pass through the join tree. From its (extended) local system of constraints, the object that started the computation may now compute the least upper bound to our probability of interest. The result obtained is the same as when obtained directly from the joint system of constraints. The intuition of this property is that when the process has again reached the object that started the computation, this object has been 'informed' of all constraints of the entire joint system. By directing the same process once more towards the root and the leaves of the tree, all objects can be brought into this state. So, in three passes through a join tree, each object locally has a kind of global knowledge concerning the joint system of constraints. It will be evident that for any probability of interest which is local to a clique we can now compute a probability interval locally.

The following algorithm describes these three passes. Without loss of generality we assume that the computation is started by the root $Cl_s$ of the clique tree $T_G$. It performs the following actions:

1. Send a request for information to all neighbours and wait.

2. If a return message, having the form of a system of constraints, has been received from *all* neighbours, then add these systems of constraints to the local system of constraints $D_s x_s = m_s$, $x_s \geq 0$; compute the feasible set $F_s$ of the resulting system and derive from it the (convex) set $\{T_{s,j} x_s | x_s \in F_s\}$, for each neighbour $Cl_j$ of $Cl_s$.

3. For each such neighbouring clique $Cl_j$, send this information as a system of constraints to $Cl_j$ using $T_{s,j} x_s - T_{j,s} x_j = 0$.

Each leaf $Cl_i$ of $T_G$ performs the following actions:

1. Wait for a message.

2. If a request for information is received, then compute the feasible set $F_i$ of the local system of constraints $D_i x_i = m_i$, $x_i \geq 0$, and derive from it the set $\{T_{i,j} x_i | x_i \in F_i\}$, for the neighbour $Cl_j$.

3. Send this information as a system of constraints to $Cl_j$ using $T_{i,j} x_i - T_{j,i} x_j = 0$, and then wait for a message.

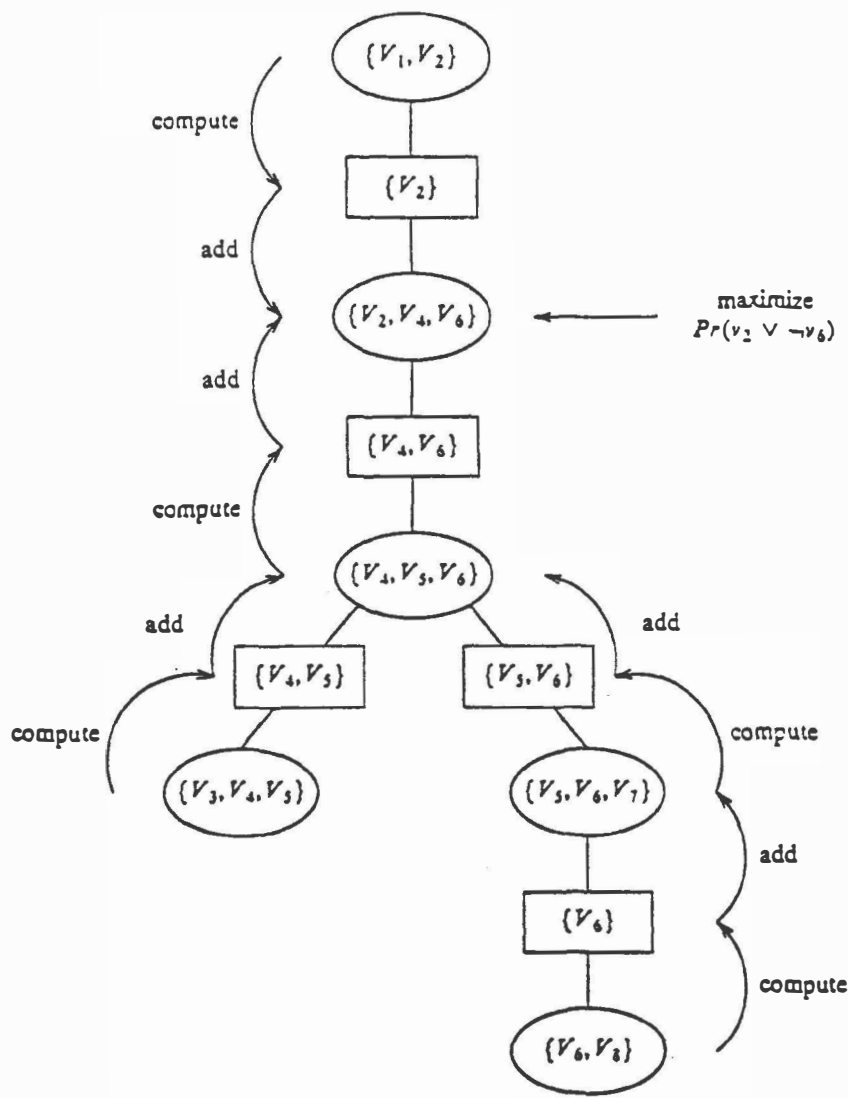

Figure 2

4. If a system of linear constraints is received, then add this system to the local system of constraints $D_i x_i = m_i$, $x_i \geq 0$.

For each interior vertex $Cl_i$, let the vertex $Cl_i^+$ be defined as the vertex on the path from $Cl_i$ to $Cl_s$, and let $Cl_i^-$ be defined as the set of all other neighbours of $Cl_i$. Each interior vertex $Cl_i$ performs the following actions:

1. Wait for a message.

2. If a request for information is received from the neighbour $Cl_i^+$, then pass this message on to all other neighbours $Cl_j \in Cl_i^-$.

3. If systems of constraints have been received from all neighbours $Cl_k \in Cl_i^-$ (or from $Cl_i^+$, respectively), then add these additional systems of constraints to $D_i x_i = m_i$, $x_i \geq 0$; compute the feasible set $F_i$ of the resulting system of constraints and derive from it the set $\{T_{i,j} x_i | x_i \in F_i\}$ for $Cl_j = Cl_i^+$ (or for each $Cl_j \in Cl_i^-$, respectively).

4. For each such clique $Cl_j$, send this information as a system of constraints to $Cl_j$ using $T_{i,j} x_i - T_{j,i} x_j = 0$.

The correctness of the algorithm has been proven in [Gaag, 1990]. In general, the algorithm may take exponential time. However, if the maximal clique size is small compared to the number of statistical vari-ables, the algorithm will take polynomial time. An important question for the algorithm to be of practical use is the question whether it is likely that the mentioned restriction will be met in practice. Concerning this, J. Pearl argues that sparse, irregular graphs are generally appropriate in practical applications, [Pearl, 1988b].

The probability intervals obtained after application of the algorithm may be rather wide, in fact they may be too wide for practical purposes. However, the intervals are precise and in a sense 'honest': they just reflect the lack of knowledge concerning the joint probability distribution.

## 4 Conclusion

In this paper we have presented a method for computing intervals for probabilities of interest from a partial specification of an unknown joint probability distribution. In our method we are able to take independency relationships between statistical variables into account by exploiting the topological properties of an I-map of the unknown joint probability distribution.

We conclude this paper with a brief sketch of an application of our method. The last few years, several probabilistic methods for reasoning with uncertainty have been proposed each departing from a so-called belief network; see for example the work by J. Pearl, [Pearl, 1988b], and the work by S.L. Lauritzen and D.J. Spiegelhalter, [Lauritzen & Spiegelhalter, 1988]. At present, such models are not capable of dealing with a partial specification of a joint probability distribution: in the models presented so far the belief network has to be fully quantified, that is, the initially assessed local probabilities have to define a unique joint probability distribution on the statistical variables discerned. Several contributors to the discussion of [Lauritzen & Spiegelhalter, 1988] have called attention to the difficulty of assessing all probabilities required. In the same discussion, D. Dubois and H. Prade furthermore argue that the requirement for a unique joint probability distribution on the statistical variables almost inevitably leads to replacing missing information by strong default assumptions concerning independency relationships between the statistical variables in order to be able to guarantee uniqueness.

The method presented in this paper can be used for stepwise quantifying a belief network. Departing from a partial specification of a joint probability distribution assessed by a domain expert, we can compute intervals for the 'missing' probabilities. These intervals can guide the expert in providing further information.